\documentclass{article}

\usepackage[final]{cml4impact_2022} % camera-ready
\usepackage[utf8]{inputenc} % allow utf-8 input
\usepackage[T1]{fontenc}    % use 8-bit T1 fonts
\usepackage{hyperref}       % hyperlinks
\usepackage{url}            % simple URL typesetting
\usepackage{booktabs}       % professional-quality tables
\usepackage{amsfonts}       % blackboard math symbols
\usepackage{nicefrac}       % compact symbols for 1/2, etc.
\usepackage{xcolor}         % colors
\usepackage{graphicx}       % images
\usepackage{wrapfig}        % wrap figures in text
\usepackage{amsmath}        % multiline equations
\usepackage{enumitem}       % itemize
\PassOptionsToPackage{sort}{natbib}
\usepackage{color}

\usepackage{todonotes}

\newcommand{\vin}{\ensuremath{v_{\text{in}}}}
\newcommand{\vout}{\ensuremath{v_{\text{out}}}}

\newcommand{\shrink}{\vspace{-2mm}}

\title{Neural Bayesian Network Understudy}
\shrink
\author{
  Paloma Rabaey, Cedric De Boom, Thomas Demeester \\
  IDLab, Dept.~of Information Technology \\
  Ghent University - imec \\
  Ghent, Belgium \\
  \texttt{first.last@ugent.be} \\
}

\begin{document}

\maketitle

\shrink
\begin{abstract}

Bayesian Networks may be appealing for clinical decision-making due to their inclusion of causal knowledge, but their practical adoption remains limited as a result of their inability to deal with unstructured data. While neural networks do not have this limitation, they are not interpretable and are inherently unable to deal with causal structure in the input space. Our goal is to build neural networks that combine the advantages of both approaches. Motivated by the perspective to inject causal knowledge while training such neural networks, % on unstructured data,
this work presents initial steps in that direction. %, limited to discrete variables. 
We demonstrate how a neural network can be trained to output conditional probabilities, providing approximately the same functionality as a Bayesian Network. Additionally, we propose two training strategies that allow encoding the independence relations inferred from a given causal structure into the neural network. We present initial results in a proof-of-concept setting, showing that the neural model acts as an understudy to its Bayesian Network counterpart, approximating its probabilistic and causal properties.

% [OLD]
% This work considers the problem of learning a neural network model on observational sample data, 
% \cdrc{Ik mis een motivatie en context. Waarom zou je dit willen doen? Staat best bovenaan de abstract.}
% while also encoding known causal relationships between the data variables. Although motivated by the more ambitious perspective to inject causal knowledge while training neural models on unstructured data,
% \cdrc{Zou niet met het negatieve beginnen (we wilden dit doen, maar bon, we zijn er niet geraakt, dus geven we jullie maar tot waar we geraakt zijn...)}
% this work presents initial steps in that direction, limited to discrete variables. In particular, we demonstrate how a generic neural network can be trained as a generative model based on data observations
% \cdrc{Dit is wel heel algemeen, niet? Misschien gewoon weglaten, en enkel schrijven dat je de knowledge uit een DAG encodeerd *door het trainen van een generatief neural network*?}
% as well as the knowledge of a directed acyclic graph (DAG) representing the available causal knowledge. We propose two training strategies that allow encoding the independence relations inferred from the DAG into the neural network, and present initial results in a proof-of-concept setting, showing that the neural model acts as an understudy to its Bayesian Network counterpart, approximating its probabilistic and causal properties.
\end{abstract}

\section{Introduction}
Bayesian Networks (BNs) have been the subject of a large body of research, and are considered an important technology for high-stakes decision making, in particular in the healthcare domain. \cite{Kyrimi2021review} provide a clear overview of the key properties that make BNs highly appealing in that context: (i) their ability to model complex problems involving causal dependencies as well as uncertainty, (ii) their ability to combine data and expert knowledge, (iii) their interpretable graphical structure, and (iv) their ability to model interventions and reason diagnostically as well as prognostically.  Despite efforts from the research community to support healthcare practitioners in using BNs \citep{AroraBoyne2019BNriskprediction, Kyrimi2020Medicalidioms}, their practical adoption remains very limited.  The analysis of \cite{Kyrimi2021BNadoption} indicates data inadequacies as one of several key limiting factors: BNs have a limited capacity to address continuous data, and cannot directly deal with unstructured data such as images or free text. According to \cite{unstructured_EHR}, the value of information entailed by unstructured text fields in electronic medical records for clinical decision support cannot be underestimated.

%\cdrc{Is dat zo? Waarom zou een BN slechter om kunnen gaan met continue data? Als je de distributies kan modelleren, lukt dat toch?}. 
%\thms{klopt dat je een distr moet modelleren, dus je moet al aannames maken.  Is letterlijke citatie uit die gerefereerde paper, maar 'k denk dat ze dit ermee bedoelen}
%\cdrc{Dat argument kan je ook gebruiken voor discrete data? Of daar kan je tellen allicht.}\note{daarvoor moet je geen aannames maken qua distributie}
%\cdrc{OK!}

In contrast to Bayesian networks, neural networks are well-suited for dealing with unstructured data. However, they are inherently not interpretable, which stands in the way of clinical adoption \citep{AI_CDS_survey, Peiffer2020review}. Whereas \cite{Hinton2018JAMA} argues strongly in favour of the adoption of neural networks in healthcare, he also recognises interpretability as a property that is desired by practitioners. 

Our envisioned research goal is to build models that combine the advantages of both approaches: a fully neural model that can be trained on observations including unstructured data, but that also adopts the desirable properties of BNs as listed above. Realistically speaking, we can at best expect only approximate probabilistic reasoning capabilities from a generic neural network, and it will lack the direct interpretability of a BN. However, if the model displays the causal properties of a BN, the ideas of counterfactual explanation \citep{wachter2017counterfactual} can be applied to explain its predictions. %, which is typically not possible standard, discriminatively trained neural networks.

This paper forms the first step along that outlined research track. We focus on the problem of training a neural network on a given set of discrete data samples as well as knowledge of a Directed Acyclic Graph (DAG) \citep{AI_russell_norvig} reflecting causal relationships between the variables. We do not yet aim to augment such models with unstructured data, but for now require that it behaves as a BN with the given DAG, if we were to train it on the same samples. In this sense, our model can be considered a \emph{neural understudy} to the BN. We want to emphasise that our current preliminary work does not aim to build a neural network which outperforms its Bayesian network counterpart, but one that approximates its capabilities. After this necessary first step, we expect to reap the benefits of the neural network understudy by incorporating unstructured data. While the current work presents results only in a proof-of-concept setting, the possibility to reason under uncertainty, while combining causal structure and unstructured data, has great potential in real-world (clinical) applications.

In short, this paper makes the following contributions: 
(1) we propose a neural network model able to infer the probability of a set of target variables conditioned on observed evidence,
(2) we introduce two approaches for injecting DAG knowledge while training the neural network, and
(3) we present empirical results (including robustness analysis) in a small proof-of-concept setting. The necessary code to reproduce these results can be found in our Github repository.\footnote{\url{https://github.com/prabaey/NBN-understudy}}

\shrink

\section{Related work}
\shrink

This section focuses on recent works most related to the proposed ideas; we do not aim to provide a broad overview in terms of related research fields. 
%In this section, we do not aim to provide an overview on the domain of causal machine learning and related research tracks. Instead, we choose to only address works that strongly relate to our proposed idea. 
%
A basic premise for the presented work is the existence of causal knowledge in addition to observational data, to be injected into a neural network model. Our goal is therefore not \emph{causal discovery}, intended to recover the causal mechanisms underlying the distribution from which the observed data was generated \citep{learning_neural_causal_models, differentiable_causal_discovery}. 
%Instead, we assume to have information on the causal structure and attempt to inject this information into a neural network.

\cite{inducing_causal_structure} present a method to align different parts of a neural network with nodes in a causal graph, 
%elements of causal knowledge a proposed high-level causal mechanism which may guide the task, 
resulting in a model with improved interpretability. \cite{causal_learning_explanation} pursue a similar goal, summarising a neural model into a Bayesian causal model, to provide counterfactual explanations for the neural model. Instead of starting from a neural network and using causal models to understand or guide its inner workings, we work the other way round.

In \cite{constructing_deep_NNs}, conditional independencies in the input distribution are encoded hierarchically in the network structure, as a way of performing unsupervised neural network structure learning. \cite{deep_structural_causal_models} and \cite{deep_causal_graphs} present each parent-child relation in a given structural causal model as a deep neural network unit. While we don't explicitly align parts of the neural network with a causal model, we do include structural information in the training process as a regulariser. This is related to the work from \cite{CASTLE}, who propose a prediction task with a regularisation objective to detect causal relationships between features. It is not based on prior knowledge of causal relationships, as in our work.

Attempts have been made to build neural networks that implement belief propagation \citep{pearl_probabilistic_reasoning} to improve on accuracy and efficiency \citep{neural_enhanced_belief_propagation, belief_propagation_NN}. Our proposed neural networks also aim at answering probabilistic queries, but can freely learn the inference mechanics from the data.

Another key related contribution is DeepProbLog \citep{Manhaeve18_deepproblog}. This neuro-symbolic model is able to implement our envisioned goal: its probabilistic logic programming core can represent any BN exactly, and its so-called neural predicates are able to encode unstructured data. However, whereas a fully neural model can at best reason only approximately with probabilities, it may have some benefits over the exact probabilistic reasoning component in DeepProbLog. Indeed, in future research we aim to model complex realistic observational data. Some of its properties a suitable model should be able to capture, may be hard to express in a probabilistic program.

Finally, our work is not to be confused with the so-called \emph{Bayesian neural networks} \cite{bayesian_neural_networks}, aiming to obtain a probability distribution over neural networks. Instead, our goal is training a neural understudy for a Bayesian Network, i.e., a neural model that behaves similarly.

\shrink
\section{Training a neural understudy of a Bayesian network} % methods

The following paragraphs describe our proposed neural architecture with full flexibility in the choice of input and output variables (Section~\ref{subsec:neuralbaseline}), followed by our proposed training strategies to inject causal structure into the model (Section~\ref{subsec:independencetraining}). 

\subsection{Neural architecture and training}\label{subsec:neuralbaseline}

A fully specified BN can be queried by providing inputs to any selection of variables (henceforth called the \emph{evidence}), after which the probabilities for each of the remaining variables (the \emph{targets}) can be inferred, conditioned on the evidence. We will require the same ability from our neural counterpart, and show that this can be achieved by using complementary masks at its inputs and outputs. This only induces restrictions on the input and output layers but not on the internal neural architecture. % \cdrc{Moet laatste zin er al bij?}

We focus on problems involving a set of $N$ discrete variables $\mathcal{V} = \{X_1, X_2,..., X_N\}$, in which $X_i$ can take $n_i$ possible values in $\mathcal{X}_i = \{x_{i1},...,x_{in_i}\}$.
%in which $X_i$ can only take $n_i$ possible values, $X_i = x_i$, whereby $x_i \in \mathcal{C}_i = \{c_{i1},...,c_{in_{i}}\}$ 
Given a set $\mathcal{E}\subset \mathcal{V}$ of $m$ evidence variables ($m<N$), each with an assigned value $\{E_1=e_1, ..., E_m=e_m\}$ (for convenience written as $\mathcal{E}=e$), our goal is to train a neural network to predict the probability distribution $P(X\vert \mathcal{E}=e)$ for each of the remaining variables $X$. The latter are called the target variables, aggregated in the set $\mathcal{T}=\mathcal{V}\setminus \mathcal{E}$. 

\begin{figure}[t]
  \centering
  \includegraphics[width=\textwidth]{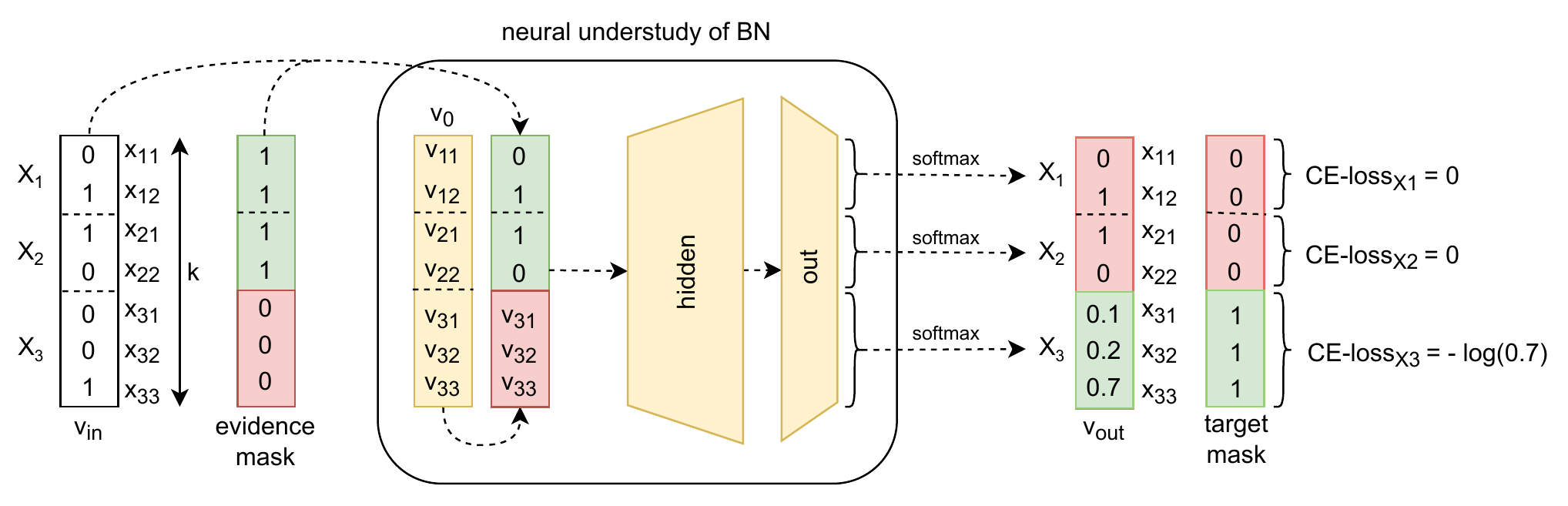}
  \caption{Illustration: training with evidence $\mathcal{E}=\{X_1,X_2\}$ and targets $\mathcal{T}=\{X_3\}$, for $P(X_3\vert X_1=x_{12}, X_2=x_{21})$. The mask at the input (output) indicates selection of evidence (target) entries.
  %To train the model, we feed it samples as they are observed in the training set and randomly mask out a target set of variables, the other values are taken as evidence. In this example, $X_1 = F$ and $X_2 = T$ is the evidence, while $X_3$ is the target. At the output of the network, only the predicted distributions of the target variables contribute to the cross-entropy loss, in which these probabilities are contrasted with their observed values at the input.
  }
  \label{fig:NN_training}
\end{figure}

After training on an observed set of individual samples of the form $\{X_1=x_1,\ldots,X_N=x_N\}$, our model should be able to deal with any selection of evidence variables. The key idea to achieve that is through dynamic masking (i.e., a different mask for any variable split $\mathcal{E}$ vs.~$\mathcal{T}$): all variables are present at the model input and output, but target variables are filtered out of the inputs through an evidence mask, whereas predictions for the evidence variables are filtered from the outputs through a target mask. This is illustrated in Fig.~\ref{fig:NN_training}.

\shrink
\paragraph{Model architecture:} Each input to the model is a vector $\vin$ of dimension $k=\sum_{i=1}^N n_i$, and is formed by concatenating the $n_i$-sized one-hot representation of the value held by each of the $N$ variables $X_i$ ($i=1,\ldots,N$). The masked positions of the target variables in $\vin$ are substituted by the corresponding components of a static vector $v_0$.
%These values are however only known for the evidence variables. For the target variables, instead, 
%the corresponding entries of a static  vector $v_0$ are used to complete $\vin$. This can be implemented conveniently through masking.
The input $\vin$ is subsequently passed through one or more hidden layers, resulting at the output in a $k$-dimensional vector of logits. These are then locally normalised through softmax activations, for each variable $X_i$ considering its corresponding $n_i$ entries. The output vector $\vout$ is finally constructed by replacing the entries at the positions of evidence variables, by the corresponding one-hot representations from $\vin$.  The values corresponding to the target variables can be interpreted as the predicted target distributions given the evidence. This is illustrated in Fig.~\ref{fig:NN_training}. Note that the $k$-dimensional vector $v_0$ is obtained by applying the linear output layer with per-variable softmax normalisation on a randomly initialised trainable vector. 

\shrink
\paragraph{Model training:} Whereas a BN is naturally able to answer any query once its conditional probability tables are specified, we need to explicitly train our neural network for that ability. To that end, we iterate over the observed samples, randomly dividing the variables into evidence set $\mathcal{E}$ and targets $\mathcal{T}$,
%This training approach can be readily modified to deal with incomplete samples as well. 
%training: CE loss + formula + explanation
%ref to proof in appendix that this leads to correct probs.
%For training on a set of samples containing observations of all variables (divided into $\mathcal{E}$ and $\mathcal{T}$), \cdrc{Voorgaande mag weg? Of herschrijven naar iets: the model is trained by minimizing the summed ...} 
and minimise the average cross-entropy loss for the target variables. %, which can be achieved by masking $\vout$ with the complementary mask as used at the input.
The training loss can be expressed as $\mathcal{L}^{T} = \frac{1}{|\mathcal{T}|} \sum_{X_i \in \mathcal{T}} -\log \hat{p}_{ij}$, in which $\hat{p}_{ij}$ denotes the entry in $\vout$ corresponding to the actual observed input value $x_{ij}$ of target variable $X_i$, i.e., the predicted output probability for the target class of $X_i$. 

Even though the model only receives discrete data samples and therefore never observes the class probabilities it is meant to predict, this training strategy steers the predicted probabilities towards the empirical (conditional) class probabilities. It only requires that each sample's frequency of occurrence during training corresponds to its relative frequency in the set of observations. Appendix \ref{app:sample_based_training} proves this for the canonical case of two binary variables, but the presented derivation can be extended to the multi-variable case. %, as confirmed in the results for our 7-variable proof-of-concept example (Section~\ref{sec:results}).

\subsection{Training with causal structure} \label{subsec:independencetraining}
We now propose the following two training strategies to augment the neural network with knowledge of the causal structure between the variables. They are based on the assumption that the causal knowledge can be represented as a DAG, reflecting the (conditional) independence relations that exist between the variables. 

\shrink
\paragraph{1. Injecting independence relations through regularisation (REG):} 
In general, we can say a given DAG implies $M$ pairwise conditional independence relations between two variables, conditioned on a set of observed variables, shortly written as $X\perp Y \mid \mathcal{A}$, with $X, Y \in \mathcal{V}$ and the conditioning set $\mathcal{A} \subset \mathcal{V} \setminus \{X, Y\}$. 
The first proposed training strategy is to inject these independence relations by constructing regularisation loss terms $\mathcal{L}_{X\perp Y \mid \mathcal{A}}^{\text{REG}}$ that quantify how strongly the model violates them:

\shrink
\begin{equation} \label{eq:IR_loss}
    \mathcal{L}_{X\perp Y \mid \mathcal{A}}^{\text{REG}} = \frac{1}{n} \sum_{j=1}^{n} \big(\hat{p}(X = x_j \mid \mathcal{A} = a, Y = y) - \hat{p}(X = x_j \mid \mathcal{A} = a, Y = y')\big)^2
\end{equation}

in which the summation runs over all $n$ possible values of $X$. The model's predicted probability for $X=x_j$ with as evidence the assignment $\mathcal{A}=a$ and $Y=y$, is denoted as $\hat{p}(X = x_j \mid \mathcal{A} = a, Y = y)$.
The loss term expresses that if the independence relation were satisfied by the model, its predicted probability for any value of $X$, given any evidence $\mathcal{A}=a$, should be independent of the value assumed by $Y$. The conditioning set $\mathcal{A}$ is randomly instantiated every time the corresponding loss term applies during training, and the values $y$ and $y'$ of $Y$ are chosen randomly (with $y\not = y'$). 

During training, the regular loss $\mathcal{L}^T$ per data sample is augmented with the regularisation loss $\mathcal{L}^{\text{REG}}$ of one sampled independence relation, weighted with a hyperparameter $\alpha$.

\shrink
\paragraph{2. Injecting independence relations through evidence corruption (COR):} The second training strategy is based on the intuition that for particular observed data samples, the value of some of the evidence variables may no longer matter besides the other observed variables, when accounting for the relevant independence relations. During training, we detect these cases, and randomly corrupt (i.e., re-sample) those values in the evidence presented to the network.
%The second approach applies the independence relations on the fly to corrupt the inputs encountered by the model during training. This approach integrates the two objectives: learning the conditional probabilities as reflected by the training samples, while at the same time using information from the causal structure to teach the model which variables should be conditionally independent. 

Consider training on a particular observed sample, the variables divided into evidence $\mathcal{E}$ and targets $\mathcal{T}$. We then go through the known independence relations $X \perp Y \mid \mathcal{A}$, to see which ones are relevant to the training instance. This is the case if either of the following two conditions hold: (1)  $(\{Y\} \cup \mathcal{A}) = \mathcal{E}$ and $X \subset \mathcal{T}$, or (2) $(\{X\} \cup \mathcal{A}) = \mathcal{E}$ and $Y \subset \mathcal{T}$. 
%
%Assume that during training, we have sampled a random mask which divides all variables into a target set $\mathcal{T}$ and an evidence set $\mathcal{E}$. We then try to match this input with an independence relation which applies to this combination of evidence and targets. Say we have an independence relation of the form $X \perp Y \mid \mathcal{A}$, with $X, Y \in \mathcal{V}$ and $\mathcal{A} \subset \mathcal{V} \setminus \{X, Y\}$. This IR forms a match with the input when either of the following two conditions hold: (1) $\{Y\} \cup \mathcal{A} \subset \mathcal{E}$ and $X \subset \mathcal{T}$, or (2) $\{X\} \cup \mathcal{A} \subset \mathcal{E}$ and $Y \subset \mathcal{T}$. 
%
When condition (1) holds, the predicted outcome of target $X$ should not depend on the observed value of $Y$, as prescribed by the independence relation, since the conditioning set that is needed for this relation to hold is indeed part of the evidence. The observed input value for $Y$ can hence be disregarded, and we randomly assign a new value from its possible classes. Such a corruption of the input will condition the model to ignore $Y$ when predicting $X$, given that all variables in $\mathcal{A}$ are also provided as evidence. Note that predictions for other variables in $\mathcal{T}$ need to be done based on the original (i.e., non-corrupted) value for $Y$, as there is no guarantee that the same independence relation holds for those targets as well.  We apply a similar reasoning when condition (2) holds, now corrupting the value of $X$ instead of $Y$. If no relevant independence relation is found for a given selection of evidence and targets, the input sample is not corrupted and passed as-is to the model. 
In Appendix \ref{app:IR_corruption}, we consider the basic two-variable case and show that the corruption strategy leads to the desired predicted probabilities.

\shrink
\section{Neural understudy of Bayesian network: proof-of-concept} \label{sec:results} 

The experiments presented in this section aim at answering the following research questions:
\begin{itemize}[noitemsep, leftmargin=5mm, topsep=-2pt]
    \item [-] \textbf{RQ1}: How does the basic neural network perform in comparison with a Bayesian Network trained on the same set of samples, in terms of accuracy in predicted probabilities and in terms of sample efficiency? %How does this performance relate to sample efficiency? 
    (Section \ref{sec:comparison_models})
    \item [-] \textbf{RQ2}: What is the effect of injecting causal structure knowledge into our neural network on its performance? (Section \ref{sec:comparison_models})
    \item [-] \textbf{RQ3}: How does each model's performance change when faced with a partially incorrect specification of the causal structure? (Section \ref{sec:robustness_miss_spec}) 
\end{itemize}

To this end, we compare the following models, trained on an artificial dataset of samples generated from a given joint distribution, as specified in the next section: % (see Section \ref{sec:evaluation}):
%on a small dataset.  The models under comparison are trained on the same samples,
%compares  We will now present initial results on the performance of four models for a small artificial data set. All models receive the same samples during training, 
%which are randomly generated from a given ground-truth joint distribution (see Section \ref{sec:evaluation}). 
%The models we compare are the following. 
\begin{itemize}[noitemsep, leftmargin=5mm, topsep=-2pt]
    \item [-] \textbf{Bayesian network baseline (BN)}: Bayesian Network with the correct DAG, its joint probability distribution estimated from the training samples using Maximum Likelihood Estimation with a K2 prior (see Appendix \ref{app:BN_implementation}). %It is a Bayesian Network n a Bayesian network, this comes down to estimating the distribution (in the form of conditional probability tables or CPTs) of each node given its parents \citep{AI_russell_norvig}. For more information on training and inference in our Bayesian model, we refer to Appendix \ref{app:BN_implementation}.
    %To this end, we use the maximum likelihood estimator. To answer probabilistic queries with the trained model, we use variable elimination as an inference technique.
    \item [-] \textbf{Neural network baseline (NN)}: Basic neural network approach  
    %This is the basic version of our neural network which can answer conditional probabilistic queries, 
    as presented in Section \ref{subsec:neuralbaseline}, implemented in a single-layer feed-forward model (see Appendix \ref{app:NN_training_details} for details). 
    %We choose to use a standard feed-forward neural network, see Appendix \ref{app:NN_training_details} for the exact design.%This network has no notion of the causal structure that connects the variables and can only learn from the training samples.
    \item [-] \textbf{NN with DAG-based regularisation (NN+REG)}: This model extends the basic NN model with independence relation information, through the regularisation technique REG (Section \ref{subsec:independencetraining}). %, meaning it has some notion of the true DAG structure that connects the variables. 
    %receives additional information on the causal structure, on top of the training samples. 
    \item [-] \textbf{NN with DAG-based corruption (NN+COR)}: This model, also based on NN, injects independence relations during training with the corrupted inputs strategy %, extends the basic NN model with DAG knowledge, now trained with the input corruption strategy  training  corrupted  architecture some knowledge on the DAG structure, now by applying the second technique 
    introduced in Section \ref{subsec:independencetraining}. \\ %The input samples are corrupted according to the causal structure to softly teach the model some independence relations. 
\end{itemize} 
\shrink

%First, Section \ref{sec:evaluation} will address how we evaluate the performance of our models and what dataset we use for our preliminary investigation. The two sections that follow after will answer three concrete research questions:

\shrink
\subsection{Evaluation} \label{sec:evaluation}

\begin{wrapfigure}{r}{0.35\textwidth}
  \centering
  \shrink\shrink\shrink\shrink
  \includegraphics[width=0.33\textwidth]{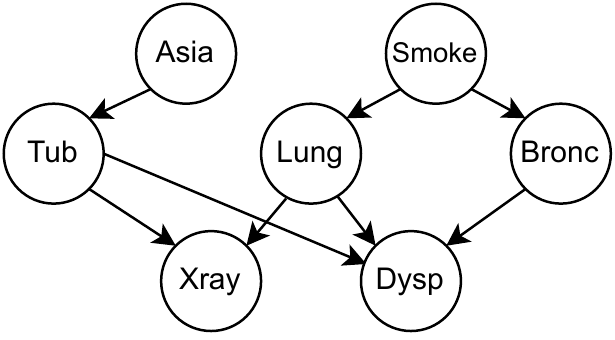}
  \caption{The ``Asia'' BN model, %based on \citep{asia}, 
  describing a basic lung cancer detection system (for details, see Appendix \ref{app:BN_implementation}).}% for the conditional probability tables (CPTs) that make up the joint distribution.}
  \label{fig:asia_model}
\end{wrapfigure}

%\thms{We construct training instances by drawing samples from a fully specified Bayesian network.} %(structure and conditional probability tables) 
%to generate artificial data samples which form a training set. 
%\thms{In particular, we use} 
%For our proof-of-concept experiments, investigation of our methods, we chose a small ground-truth Bayesian network:
For the experiments in this section, we make use of the ``Asia'' network based on \citep{asia}, as depicted in Fig.~\ref{fig:asia_model}. Despite its simplicity, with only 7 binary variables, it has
%We chose this network \thms{for our proof-of-concept experiments,} due to its simplicity: \thms{it has 7 binary nodes, but with}
%all nodes are discrete, the number of variables is small, and there is 
sufficient connectivity between the variables to make the problem non-trivial. Following the rules of D-separation \citep{d_separation}, we can in total extract 191 unique independence relations from its DAG, for the training strategies NN+REG and NN+COR. 
%The ``Asia'' DAG structure used here encodes 191 unique independence relations. 
%We would like to emphasise that we simply aim to present some preliminary results to get an idea of the performance of our proposed methods on a small artificial network. 
%The models we propose are general enough to deal with any number of variables and Bayesian network structure, but their scalability should be investigated in future work.

From the ground-truth ``Asia'' model (detailed in Appendix \ref{app:BN_implementation}), we can draw data samples to use as training instances.
%Through the ``Asia'' model, we have access to the ground-truth joint distribution which generated the samples that are fed into our models. 
%By applying the technique of 
Through variable elimination \citep{AI_russell_norvig},  %\thms{on the joint probability distribution} %in this ground-truth model, 
we can build test queries with the ground-truth conditional probabilities of target variables, for any particular assignment of evidence variables. Just like during training, this means that one query may contain multiple target variables. % which specify the conditional probability we expect when receiving a particular evidence combination. 
We can then evaluate models by iterating through the test queries, and calculating 
%To evaluate a model, we iterate through all test queries in a set and calculate 
the mean absolute error (MAE) between the predicted target distribution and the desired one. For each query, we compute the MAE per target variable, sum all their contributions and divide by the number of targets in the query. We use two different test sets. 
The first set contains all possible evidence assignments, in total 2,059 queries, and leads to the \textbf{total MAE}.
%out of the variables and their possible values, 
%measuring the \textbf{total MAE} (over 2059 queries). 
The second set contains 1,000 queries, each constructed by drawing a sample from the joint probability distribution, then randomly selecting the set of evidence variables, and obtaining the conditional probabilities for the target variables. It measures the \textbf{sample MAE}, assigning more weight to the model's performance for queries with commonly observed evidence values, while the total MAE puts equal focus on common as well as rare assignments of the evidence variables. %combinations. 
%For more information on how we build up these test sets and their respective size, we refer to appendix \ref{app:evaluation}.

\subsection{Performance of Bayesian network vs. neural understudy} \label{sec:comparison_models} 

\begin{figure}[t]
  \centering
    \includegraphics[width=0.49\textwidth]{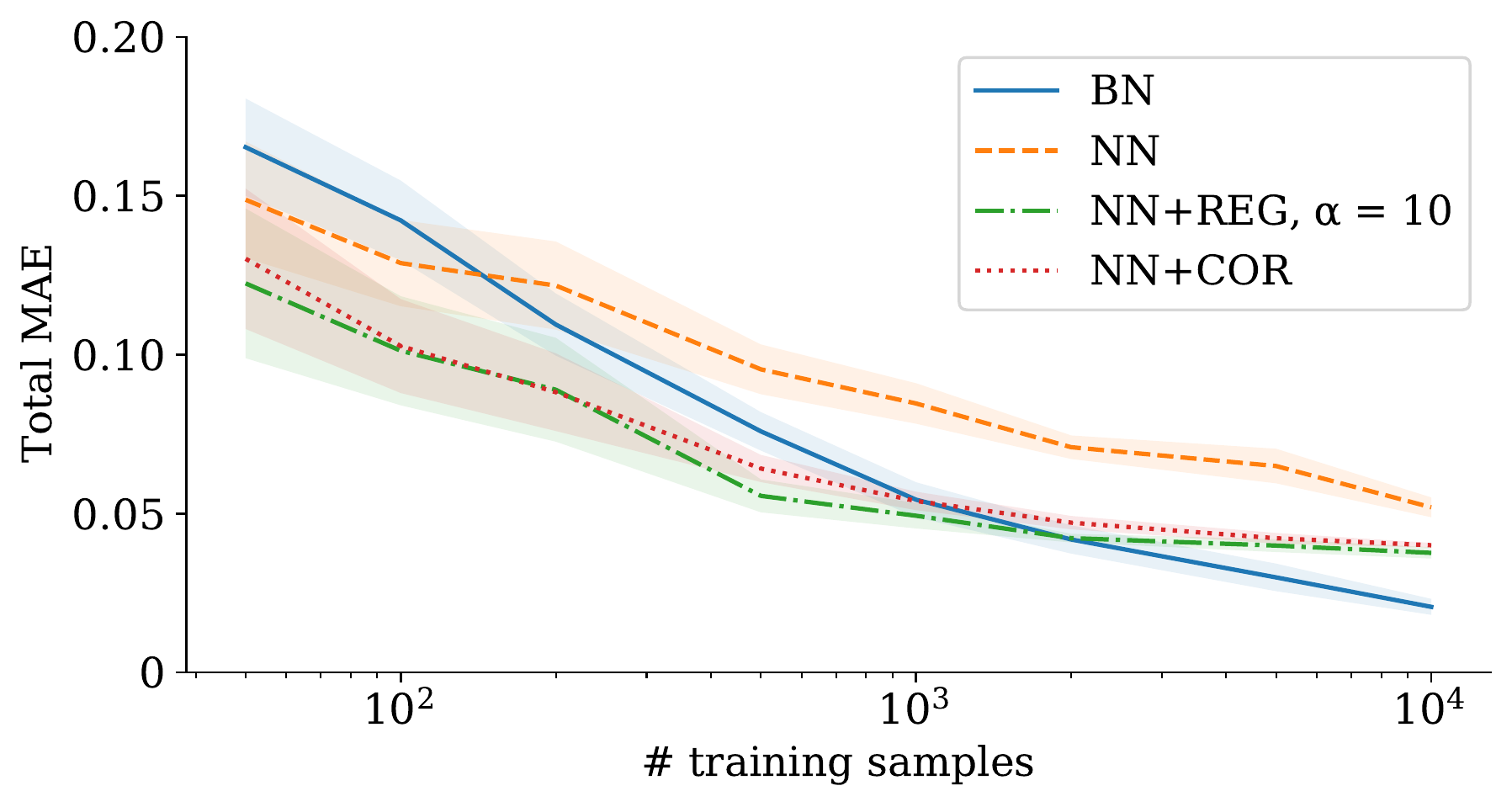}
    \includegraphics[width=0.49\textwidth]{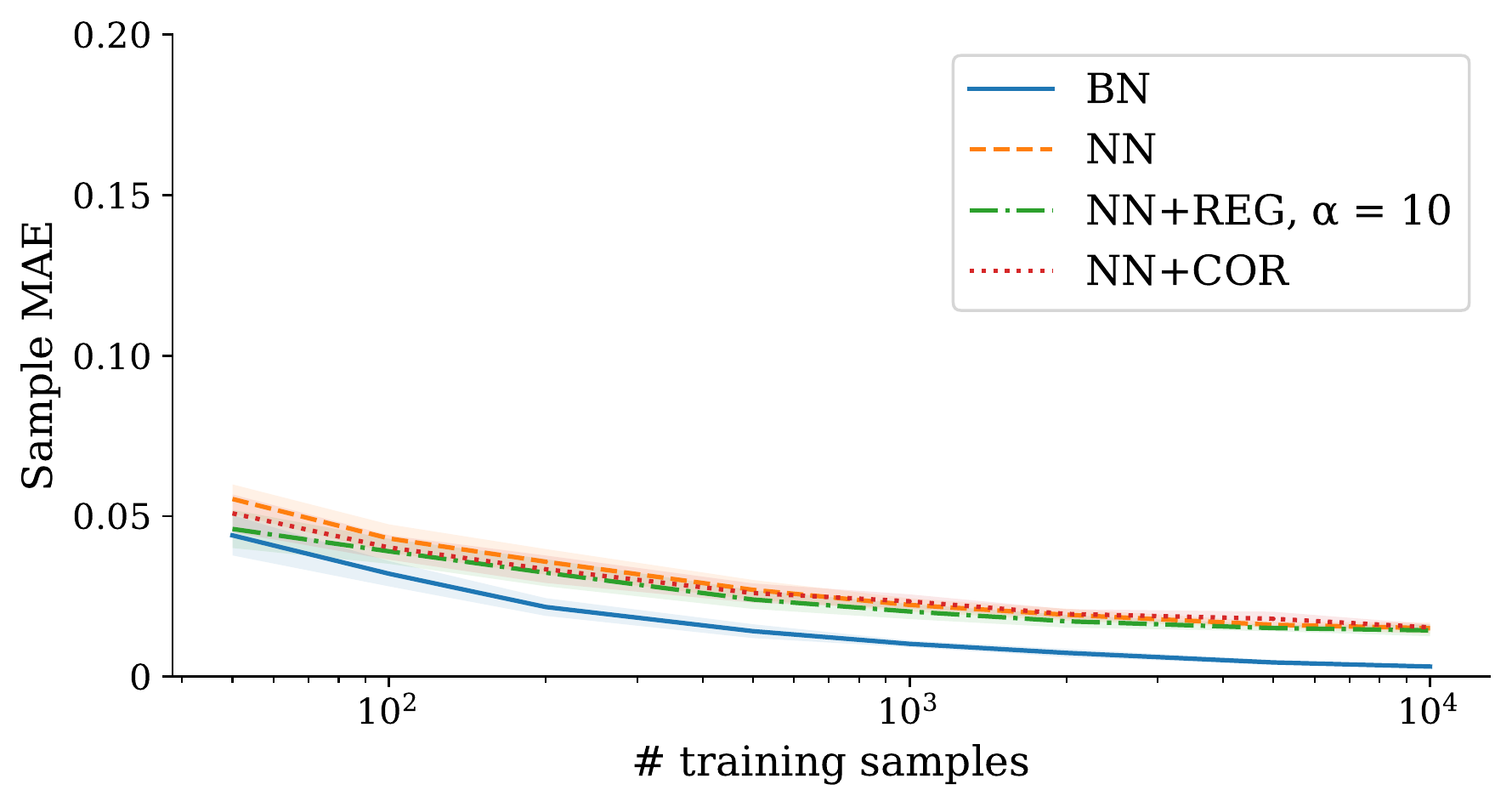}
  \caption{Comparison of 4 models in terms of total MAE and sample MAE for different sizes of the training set. %For each size, 10 different seeds are used to sample a training set from the ground-truth distribution, which are all fed to the 4 models. %These sets of samples are then fed into the 4 models. The same seeds are used for the random initialisation of the neural network. 
  The lines reflect the mean across 10 random seeds, used to sample the training set and initialise the neural networks. The shading represents the 95\% confidence interval.}
  \label{fig:performance_MAE}
\end{figure}

In Fig.~\ref{fig:performance_MAE} we compare the sample efficiency of our 4 models in terms of total and sample MAE.
%The curves depict the mean and confidence intervals for 10 different seeds used to pick the training samples, for each sample size. 
For the NN+REG model, we set the regularisation parameter $\alpha$ to 10. The training hyperparameters (hidden layer dimensions, batch size, learning rate, number of training epochs) are kept fixed for all neural models to allow for a direct comparison. Details on the training process and hyperparameters are given in %how we chose these hyperparameters, we refer to 
Appendix \ref{app:NN_training_details}. The results allow us to answer the first two research questions.

%Based on Fig.~\ref{fig:performance_MAE} we can formulate an answer to the first two research questions. 

\shrink
\paragraph{RQ1 (Sample efficiency)} %The neural network shows better sample efficiency than the BN baseline in terms of total MAE, even it its basic form. This is remarkable, since the basic NN model does not have any knowledge of the DAG structure, while the BN does. 
The proposed training strategy leads to neural understudy models that can approximately infer conditional probabilities. For smaller training sets, the basic NN model shows similar performance as the BN baseline in terms of total MAE. As the training set grows larger, the BN model's knowledge of the DAG structure allows it to significantly overtake the basic NN model in terms of performance. The BN baseline outperforms the neural models across the board when looking at sample MAE, though the gap is not large.
%The same cannot be said for the sample MAE, where the BN baseline performs better than the neural models across the board. 

\shrink
\paragraph{RQ2 (Effect of including DAG info)} Including information on the DAG structure in the neural model improves its performance, with a more visible effect on the total MAE. In terms of that metric, injecting the independence relations during training allows the neural model to outperform the BN baseline for smaller sample sets. The NN+REG model performs slightly better than the NN+COR model, albeit not significantly. We hypothesise that this stems from the REG technique systematically iterating over all conditional independence relations, whereas the COR technique only applies a relation when the randomly sampled evidence forms a match. %The improvement in terms of sample MAE after adding DAG information is not significant. 
The strength of the neural models appears to lie in their improved performance for rare evidence combinations, which are more heavily disadvantaged in smaller datasets.

%As the size of the dataset increases, the overall MAE decreases for all models, which is to be expected. First, we focus on the total MAE. The basic neural distillation performs better than the BN baseline when the models see 500 samples or less. This is remarkable, since the basic NN model does not have any knowledge of the causal structure between the variables, while the BN does. Both techniques (IRR and IRC) to enrich the NN model with information on the independence relations perform significantly better than the basic NN model. The IRR technique performs slightly better than the IRC technique. Also note that the BN baseline exhibits a wider confidence interval, indicating its performance seems more dependent on the particular training samples it encounters, while the NN models seem a little more stable.

%The sample MAE is always lower than the total MAE. This is not unexpected, since the sample MAE is a reflection of the performance of the model for queries containing more common combinations of observed evidence, while the total MAE assigns equal weight to very rare probabilistic queries. The BN baseline performs better than the NN distillation models for all sizes of the training set. We also note that the difference between the basic NN model and the version enriched with information on the IRs is not significant in terms of sample MAE. The added value of including IRs in the NN model clearly only pays off in terms of performance for evidence combinations which might barely (or not at all) appear in the training set.

\subsection{Robustness against miss-specification of causal structure} \label{sec:robustness_miss_spec}

\begin{figure}[t]
  \centering
    \includegraphics[width=0.48\textwidth]{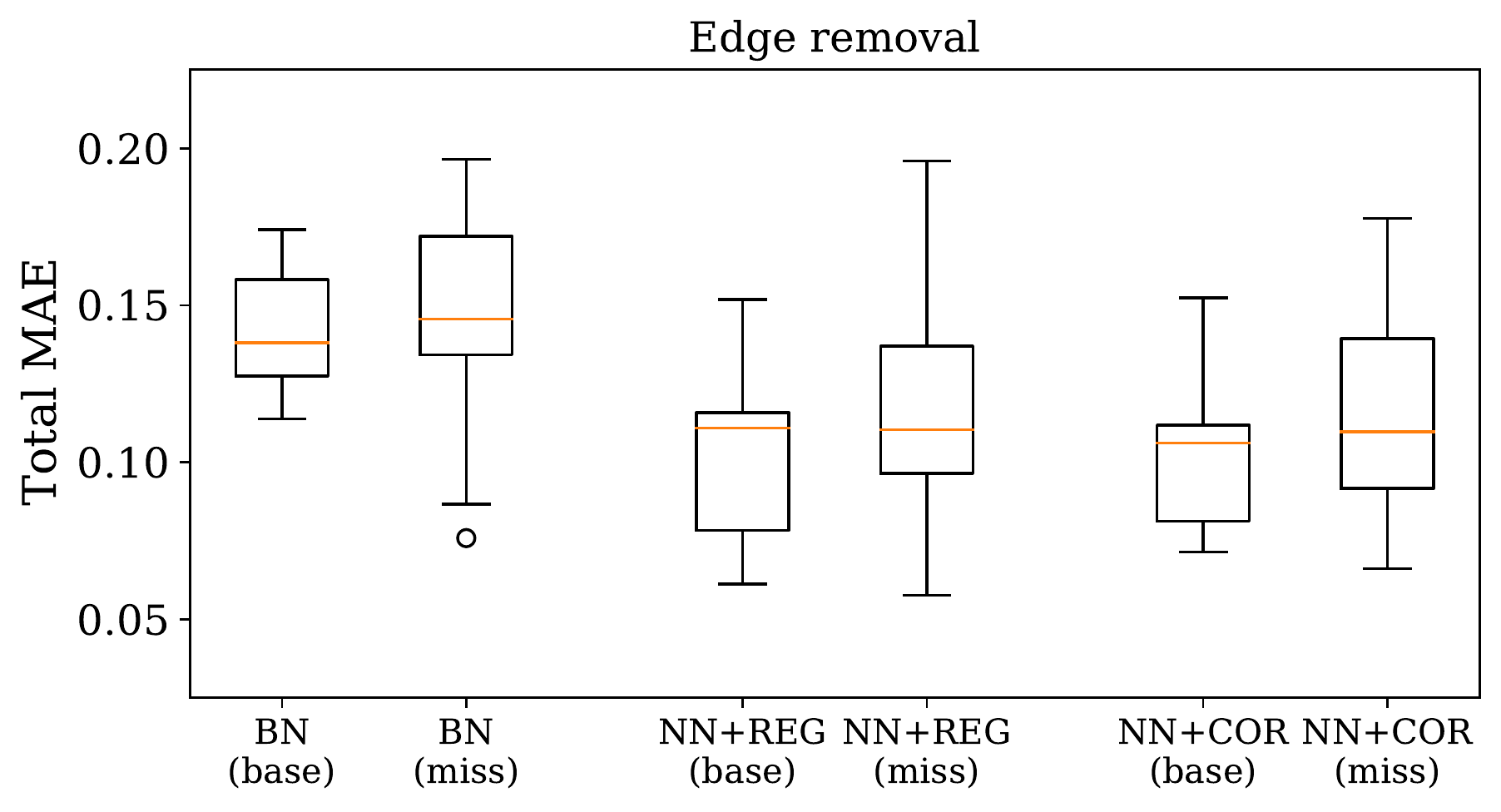}
    \includegraphics[width=0.48\textwidth]{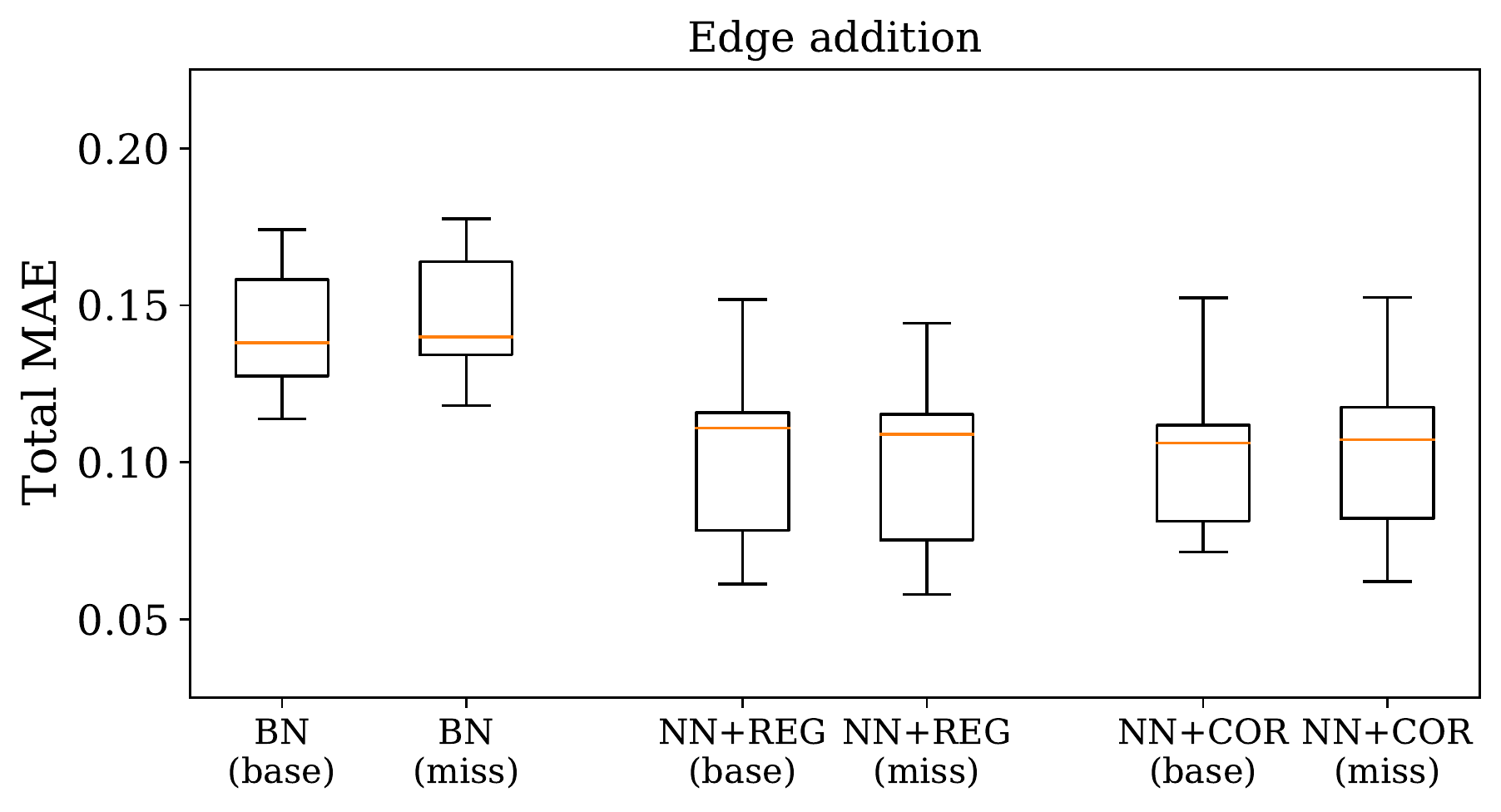}
  \caption{Boxplots visualising the total MAE for models receiving the correct DAG (``base'') vs.~a partially miss-specified DAG (``miss''), i.e., either one edge removed (left) or added (right), for different models (BN; NN+REG; NN+COR) trained on 100 observed samples. All train runs are done for 5 random seeds, and for ``miss'' runs, 5 random DAG corruptions are done (resulting in 25 runs per model for the ``miss'' setting).
  %Boxplots illustrating the effect of removing (left) or adding (right) one edge in the DAG on the total MAE. 
  %\thms{Pairwise comparison between the correct DAG (``base'') and miss-specified DAG (``miss''; one edge removed or added).} 
  %The figures show a pairwise comparison of the performance when receiving the correct DAG (``base'') and a partially miss-specified DAG (``miss''). 
  %During each run, all models receive 100 training samples and one version of the DAG. The ``base'' results are obtained by running each model with 5 sample seeds (used to sample the training data and initialise the neural networks), while the ``miss'' results are obtained over 25 runs per model type (5 DAGs with one edge removed/added, each run with the same 5 seeds). %For training, we use the same settings as before (see Section \ref{sec:comparison_models}).
  }
  \label{fig:robustness_total_MAE}
\end{figure}

%On the left, we see the effect of removing an edge for the BN, NN+IRR and NN+IRC model (the basic NN model has no notion of the independence relations and is automatically not affected by a change in causal structure). All results are obtained by training the models with 100 samples, with the same settings from Section \ref{sec:comparison_models}. The ``base'' boxplots show the distribution of MAE for the correct causal structure when 5 different seeds are used for sampling the training set and initialising the neural networks. The ``miss'' boxplots are obtained by removing one random edge at a time, and running the models with 5 seeds for each configuration (which makes 25 runs in total). On the right, we see the effect of adding one random edge at a time. Here, we were careful not to add edges which might introduce cycles in the network, as this goes against the definition of a Bayesian network. Similar plots can be found for the sample MAE in appendix \ref{app:robustness}. 

We now explore the impact of injecting partially incorrect information on the causal structure. To this end, we create 10 new DAGs, by randomly removing or adding \footnote{We were careful not to add edges which might introduce cycles in the network.} one edge at a time to the ground-truth DAG from Fig.~\ref{fig:asia_model}. While still being trained on samples from the correct ``Asia'' network, the BN, NN+REG and NN+COR models now get their conditional independence relations from a partially incorrect DAG. The results displayed in Fig.~\ref{fig:robustness_total_MAE} allow answering the third research question.

%Our NN+IRR and NN+IRC models now receive a different set of independence relations, automatically extracted from the miss-specified causal structures. Before running these experiments, we hypothesise that the neural distillation will be less affected by miss-specification of the causal structure since this information is injected in a soft manner, as opposed to the Bayesian network where this is an integral part of its definition. 

%Figure \ref{fig:robustness_total_MAE} shows the results of these robustness experiments in terms of total MAE. Based on this, we can formulate an answer to the third research question.

\shrink
\paragraph{RQ3 (Partially incorrect DAG)} While the inclusion of DAG information in the neural understudy brings clear additional benefits %over the BN baseline in terms of total MAE, this is only the case when the causal structure is accurately specified. 
in terms of total MAE compared to its most basic form, we must be careful to correctly specify this causal structure.
When we assume two variables to be independent when they are not (i.e. by removing an edge in the DAG), %the neural model sees a larger performance degradation than the Bayesian baseline, though none of the effects are significant. 
both the Bayesian baseline and the neural models become less stable, showing higher variation across sample sets.
Adding an edge in the DAG seems to do no harm, since this makes for a more conservative estimate of the DAG structure (we assume two variables to have some dependency when they do not).

%The effect of removing an edge is overall more noticeable than the effect of adding an edge. This makes sense, as removing an edge corresponds with assuming conditional independence between two variables where this is not the case. Adding an edge means we cannot be sure of the independence of two variables, though they still might be. 

%The Bayesian network performance seems less sensitive to the removal of edges than our neural distillations. Though a t-test confirms that none of the observed effects are significant, we still need to reject our hypothesis. Our proposed neural models do not show better robustness against miss-specification of the causal structure than Bayesian networks. In fact, the opposite is true when edges are removed. We can thus formulate the following answer to the final research question. 

\section{Conclusions and future work}  

We presented ideas for building neural networks that behave like Bayesian Networks. In future research we aim to combine the benefits of neural networks (i.e., encoding unstructured data) with some advantages of Bayesian networks, such as their ability to combine uncertainty with causal structure. As a first contribution in that direction, we presented a method to learn a neural network model on observational sample data, and two strategies to encode known causal relationships between the variables, by injecting independence relations. 

We tested our method on a single small-scale example, and saw that our proposed training strategy generally works: our neural understudy models (w.r.t.~to their Bayesian Network counterpart trained on the same samples) were able to make approximate predictions of conditional probabilities. %We even noticed that the neural models may on average achieve better sample efficiency (when testing on all possible queries) than the Bayesian Network. 
The inclusion of causal structure resulted in similar performance of the neural understudy compared to the BN baseline, with the neural understudy slightly outperforming the BN under low-sample regimes when testing on all possible queries. %brought additional performance benefits to the neural model, but the causal structure must be accurate. 
We saw the performance of the NN models become less stable as soon as incorrect independence relations were injected, though this behaviour was also observed for the BN.
%If the DAG we use as a basis for extracting conditional independence relations is missing an edge, the performance of our neural model may degrade to the point of performing worse than the Bayesian baseline. 

We see multiple avenues for future work. We will first extend our experiments to larger, more realistic settings, to see whether the conclusions based on our small example still hold. %, to see if our observations generalise to more extensive causal network structures. As there are more nodes and connections in the causal structure, it becomes harder to exhaustively list all conditional independence relations. The question remains whether the two DAG-based techniques we proposed still show their merit when they only receive part of the relations between the variables. 
%
%The work presented in this article was motivated by the perspective to inject causal knowledge while training neural models on \textit{unstructured} data (text or images), since this is where their strengths lie. 
%
We then aim to investigate how our models can be extended towards continuous variables as well as unstructured data, to be able to answer probabilistic queries concerning a combination of any of these inputs. For example, we envision the use of pre-trained language encoders, although that will require considerable adaptation of the straightforward model with corresponding discrete nodes at input and output.

%\plm{Reviewer: "Normally, this would have been valuable contribution, especially for a workshop, but the authors do not discuss how their work can be scaled to unstructured data (and what are the additional challenges of doing so compared with the simplified case they consider in this paper)." Extend the future work section a little bit, to include the encoder-decoder idea for unstructured data?}

%To this end, our network must integrate discrete, continuous and unstructured variables all at once, and we must train it to answer probabilistic queries concerning a combination of any of these input types. Such a neural model has the potential to be much more powerful than any Bayesian network, where it is not possible to integrate unstructured nodes with the rest of the network during inference. 

\bibliography{bibliography}

%\newpage

\appendix

\section{Appendix}

\subsection{Sample-based training of neural network} \label{app:sample_based_training}

\begin{wraptable}{R}{0.4\textwidth}
\centering
\vspace{-18pt}
\caption{Absolute frequency table showing the occurrence of $X = x$ and $Y = y$ in the hypothetical training set ($\nu = n_{00} + n_{01} + n_{10} + n_{11}$). \\}
\label{tab:train_freq}
\vspace{5pt}
\begin{tabular}{ c c | c } 
 x & y & freq. \\
 \hline
 0 & 0 & $n_{00}$ \\ 
 0 & 1 & $n_{01}$ \\ 
 1 & 0 & $n_{10}$ \\ 
 1 & 1 & $n_{11}$ \\ 
\end{tabular}
\vspace{-10pt}
\end{wraptable}

%It is not trivial to see how providing observed samples at the input of our neural network during training leads to an estimation of conditional probabilities at the output. 
We provide a small-scale proof of how optimising the cross-entropy loss for the observed samples leads to probabilistic outcomes. While the illustrative setting for which we provide the proof only concerns two binary variables, the proof can be extended to a more general setting. %one can easily see how it can be extended to a more general setting (more than two discrete variables, not necessarily binary).

Assume we have two binary variables, and observe $\nu$ samples 
%meaning we observe a sample 
$\{X = x, Y = y\}$ during every training epoch.
%during every training iteration, $\nu$ samples in total. 
The distribution of the training data is shown in Table \ref{tab:train_freq}. Our model simultaneously optimises three partial training losses, each corresponding to a possible evidence mask: $\mathcal{L}_{X \rightarrow Y}$ (target Y, evidence X), $\mathcal{L}_{Y \rightarrow X}$ (target X, evidence Y), $\mathcal{L}_{XY}$ (target X and Y, no evidence). The random evidence/target mask decides which loss is optimised during which iteration, and the sum of these three losses forms the overall training objective. We will calculate the optimum for each partial training loss and show that this leads to the desired predictions for the targets at hand. 

\shrink
\paragraph{Evidence $X$, target $Y$:} Optimising the partial loss $\mathcal{L}_{X \rightarrow Y}$ should lead us to an estimate for query $P(Y = y | X = x)$. The model returns a prediction $\hat{y}_{x \rightarrow y}$ for target $Y = y$ taking evidence $X = x$ as an input. Taking into account the frequencies of each sample as listed in Table \ref{tab:train_freq}, we obtain the partial loss $\mathcal{L}_{X \rightarrow Y}$ as shown in eq.~(\ref{eq:loss_ev_X_tar_Y}). Here, we use the definition of binary cross-entropy loss as listed in eq.~(\ref{eq:binary_CE_loss}) and simplify by filling in the possible values (0 or 1) for target y. 

\begin{equation} \label{eq:binary_CE_loss}
    \mathcal{L_{CE}} = - y log(\hat{y}) - (1-y) log(1-\hat{y})
\end{equation}

\begin{equation} \label{eq:loss_ev_X_tar_Y}
\begin{split}
    \mathcal{L}_{X \rightarrow Y} & = - n_{01} log(\hat{y}_{0 \rightarrow 1}) - n_{00} log(\hat{y}_{0 \rightarrow 0}) - n_{11} log(\hat{y}_{1 \rightarrow 1}) - n_{10} log(\hat{y}_{1 \rightarrow 0}) \\ 
    & = - n_{01} log(\hat{y}_{0 \rightarrow 1}) - n_{00} log(1-\hat{y}_{0 \rightarrow 1}) - n_{11} log(\hat{y}_{1 \rightarrow 1}) - n_{10} log(1-\hat{y}_{1 \rightarrow 1})
\end{split}
\end{equation}

When given $X = 0$ as evidence, we will optimise this loss for $\hat{y}_{0 \rightarrow 1}$. Equation (\ref{eq:opt_loss_ev_X_tar_Y}) illustrates how calculating the partial derivative and setting it to zero, leads to the optimum for $\hat{y}_{0 \rightarrow 1}$ (whereby $\hat{y}_{0 \rightarrow 0}$ = 1 - $\hat{y}_{0 \rightarrow 1}$). A similar derivation for $X = 1$ leads to the optimum for $\hat{y}_{1 \rightarrow 1}$ (and $\hat{y}_{1 \rightarrow 0}$ = 1 - $\hat{y}_{1 \rightarrow 1}$). In other words, the predicted value for target Y moves towards its relative frequency in the training set, conditioned on the observed evidence values. 

\begin{equation} \label{eq:opt_loss_ev_X_tar_Y}
\begin{split}
    \frac{\partial\mathcal{L}_{X \rightarrow Y}}{\partial \hat{y}_{0 \rightarrow 1}} = 0 \Rightarrow \hat{y}_{0 \rightarrow 1} = \frac{n_{01}}{n_{00}+n_{01}} ;\quad
    \frac{\partial\mathcal{L}_{X \rightarrow Y}}{\partial \hat{y}_{1 \rightarrow 1}} = 0 \Rightarrow \hat{y}_{1 \rightarrow 1} = \frac{n_{11}}{n_{10}+n_{11}}
\end{split}
\end{equation}

\shrink
\paragraph{Evidence $Y$, target $X$:} Optimising the partial loss $\mathcal{L}_{Y \rightarrow X}$ should lead us to an estimate for query $P(X = x | Y = y)$. The derivation is  symmetrical to the previous case, with the roles of $X$ and $Y$ switched. 

\shrink
\paragraph{No evidence, targets $X$ and $Y$:} By optimising $\mathcal{L}_{XY}$, the model jointly optimises its predicted outcome for queries $P(X = x)$ and $P(Y = y)$. A training case only contributes to this partial loss when the evidence mask chooses both variables as targets. Now, we can write the loss as in eq.~(\ref{eq:loss_tar_XY}), where $\hat{x}$ is the prediction for $X = 1$ given no evidence, and analogous for $\hat{y}$. We use the same notation for the relative frequencies as before, and again use the definition of binary CE-loss from eq.~(\ref{eq:binary_CE_loss}).

\begin{equation} \label{eq:loss_tar_XY}
\begin{split}
    \mathcal{L}_{XY} = 
    - (n_{10} + n_{11}) log(\hat{x}) - (n_{00} + n_{01}) log(1-\hat{x}) \\
    - (n_{01} + n_{11}) log(\hat{y}) - (n_{00} + n_{10}) log(1-\hat{y}) 
\end{split}
\end{equation}

Again, we can optimise the loss above for $\hat{x}$ and $\hat{y}$. This leads to the optima shown in eq.~(\ref{eq:opt_loss_tar_XY}). The optimum for $\hat{x}$ simply corresponds to the relative frequency of seeing $X = 1$ in the training set, which is indeed what we want as a prediction for $P(X = 1)$. The same goes for $\hat{y}$. 

\begin{equation} \label{eq:opt_loss_tar_XY}
\begin{split}
    \frac{\partial\mathcal{L}_{XY}}{\partial \hat{x}} = 0 \Rightarrow \hat{x} = \frac{n_{10}+n_{11}}{n_{00}+n_{01}+n_{10}+n_{11}} ;
    \frac{\partial\mathcal{L}_{XY}}{\partial \hat{y}} = 0 \Rightarrow \hat{y} = \frac{n_{01}+n_{11}}{n_{00}+n_{01}+n_{10}+n_{11}} \\
\end{split}
\end{equation}

We emphasise that the results above only hold if training happens uniformly over the available training instances, so that the frequency of occurrence of a target given a particular evidence actually corresponds with the empirical probability in the training data. We ensure that this is the case by properly shuffling our batches within each epoch. 

\subsection{Injecting independence relations through evidence corruption} \label{app:IR_corruption}

%To show why a corruption of the inputs according to the conditional independence relations leads to a model whose probabilistic outcomes reflect these independence rules, 
We again consider the simple setting of two binary variables $X$ and $Y$, to show that the NN+COR method works as expected. We now add the knowledge that $X$ and $Y$ are independent. The total loss is made up of $\mathcal{L}_{X \rightarrow Y}$, $\mathcal{L}_{Y \rightarrow X}$ and $\mathcal{L}_{XY}$, as defined in Section \ref{app:sample_based_training}. Since the DAG-based corruption is only executed when the evidence set is not empty, only the partial losses $\mathcal{L}_{X \rightarrow Y}$ and $\mathcal{L}_{Y \rightarrow X}$ are affected. We zoom in on how to adapt the first one according to this new setting and how this affects the predicted outputs. The derivation for the other partial loss is symmetrical. 

Say we receive a sample $\{X = x, Y = y\}$ during training and the mask indicates that $X$ is evidence, while $Y$ is the target. Since $X \perp Y$, the value of the target $y$ should be independent of the observed evidence value $x$. Therefore, we corrupt the value of $x$, setting it to 1 with probability $\gamma$ and to 0 with probability $1-\gamma$. We can use the frequencies from Table \ref{tab:train_freq} to write out the contribution of each training sample to the partial loss $\mathcal{L}_{X \rightarrow Y}$, taking into account that the evidence is corrupted for a fraction of the training samples. This is depicted in eq.~(\ref{eq:loss_IRC}). The predicted targets $\hat{y}_{0 \rightarrow 1}$ and $\hat{y}_{1 \rightarrow 1}$ are defined in the same way as described in Section \ref{app:sample_based_training}. 

\begin{equation} \label{eq:loss_IRC}
\begin{split}
    \mathcal{L}_{X \rightarrow Y} = & - (1-\gamma) (n_{01} + n_{11}) log(\hat{y}_{0 \rightarrow 1}) - (1-\gamma) (n_{00}+n_{10}) log(1-\hat{y}_{0 \rightarrow 1}) \\ 
    & - \gamma (n_{11}+n_{01}) log(\hat{y}_{1 \rightarrow 1}) - \gamma (n_{10}+n_{00}) log(1-\hat{y}_{1 \rightarrow 1})
\end{split}
\end{equation}

When taking the partial derivative of the loss as shown in eq.~(\ref{eq:opt_IRC}), we get the optima for $Y = 1$ with either value of $X$ as evidence. Due to the corruptions we implemented in the training process, we now get $\hat{y}_{0 \rightarrow 1}$ = $\hat{y}_{1 \rightarrow 1}$. The prediction for $Y$ is indeed independent of the value of $X$ (in accordance to $X \perp Y$) and simply equal to the relative frequency of observing $Y = 1$ in the training set. 

\begin{equation} \label{eq:opt_IRC}
\begin{split}
    \frac{\partial\mathcal{L}_{X \rightarrow Y}}{\partial \hat{y}_{0 \rightarrow 1}} = 0 \Rightarrow \hat{y}_{0 \rightarrow 1} = \frac{n_{01}+n_{11}}{n_{00}+n_{01}+n_{10}+n_{11}} \\
    \frac{\partial\mathcal{L}_{X \rightarrow Y}}{\partial \hat{y}_{1 \rightarrow 1}} = 0 \Rightarrow \hat{y}_{1 \rightarrow 1} = \frac{n_{01}+n_{11}}{n_{00}+n_{01}+n_{10}+n_{11}} 
\end{split}
\end{equation}

Note that the parameter $\gamma$ plays no role in the optimum for $\hat{y}$. It does not matter according to which distribution we corrupt the evidence. In our implementation, we sample uniformly over all possible classes for the variable in question to corrupt its value. We could also opt to pull a random sample from the training set and switch out the value of the evidence variable to its value in this sample.

\subsection{Bayesian network implementation} \label{app:BN_implementation}

\begin{figure}[t]
  \centering
    \includegraphics[width=0.7\textwidth]{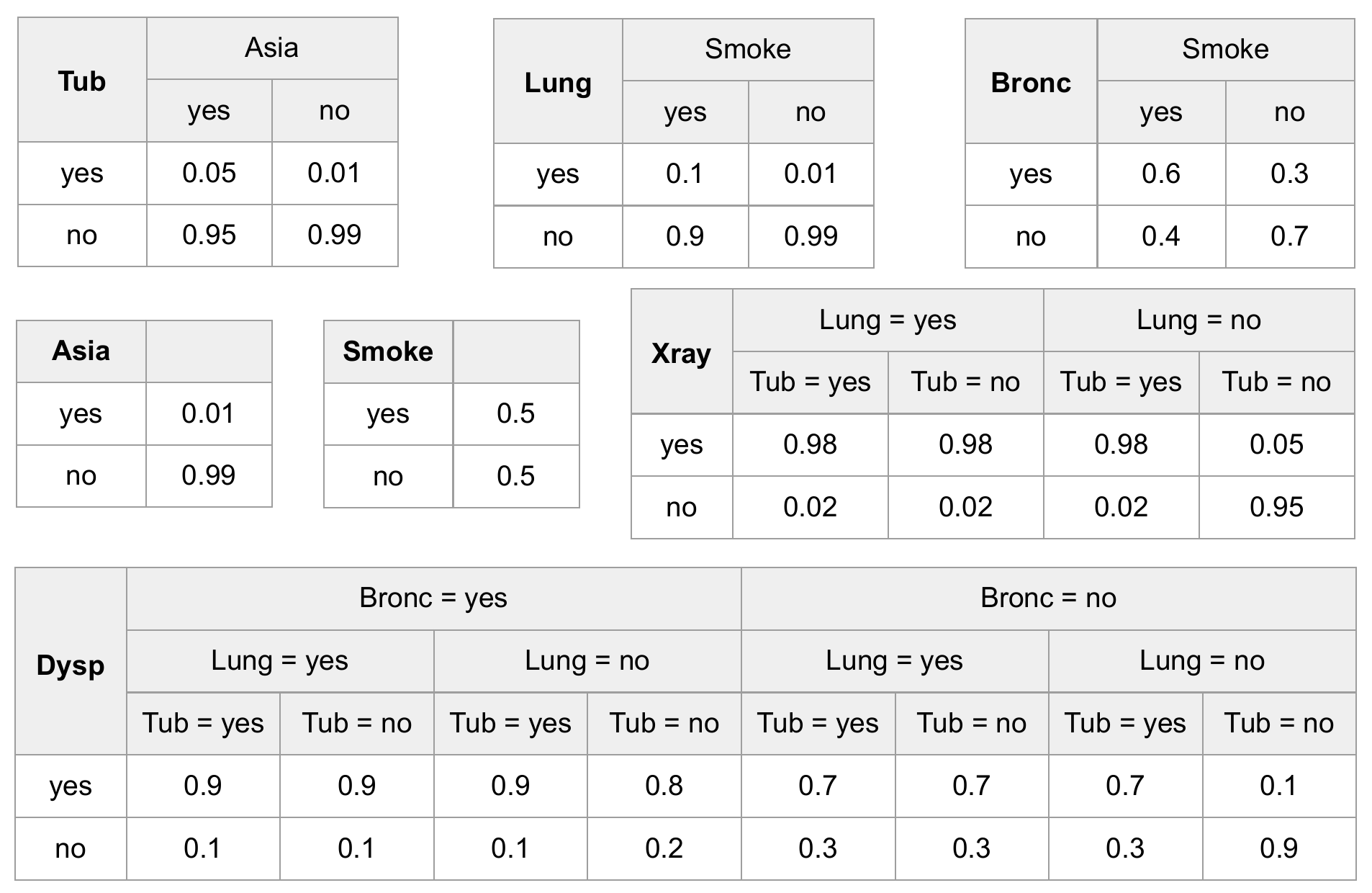}
  \caption{Conditional probability tables defining the ground-truth ``Asia'' Bayesian network, based on \citep{asia}.}
  \label{fig:asia_CPTs}
\end{figure}

We use the \textit{pgmpy} Python library \citep{pgmpy} (version 0.1.17) for sampling, training and inference in our Bayesian networks.

To train our Bayesian networks from observed samples, we use the Maximum Likelihood Estimator. This estimator studies the co-occurrence of particular values of each variable and its parents in the training set, filling up the CPTs as such. We use a K2 prior as a smoothing strategy, to counteract the extremely skewed probability distributions that might appear in the CPTs when particular combinations of variables are never observed in the training set. %This comes down to learning all conditional probability distributions $P(X_i \mid Parents(X_i))$, which are represented by a conditional probability table (CPT), when the DAG structure is given. The \textit{MaximumLikelihoodEstimator} does this by studying the co-occurrence of particular values of each variable and its parents from a set of observed samples.

Artificial samples are generated from the ground-truth Bayesian network using the method of forward sampling with a particular seed. As a ground-truth Bayesian model, from which our artificial training and test sets are created, we use the ``Asia'' model (see Fig.~\ref{fig:asia_model} for its DAG structure). The CPTs that define its joint distribution with 20 parameters are shown in Fig.~\ref{fig:asia_CPTs}.
%\textit{BayesianModelSampling} class. We always define a seed to ensure reproducibility. As a ground-truth Bayesian model, from which our artificial training and test sets are created, we use the ``Asia'' model (see figure \ref{fig:asia_model}). Usually, an additional ``either'' node is included as a child of the ``tub'' and ``lung'' nodes to reduce the number of parameters needed to specify the joint probability. We used a version of the network without this node, altering the CPTs accordingly so the joint probability distribution was not impacted. The resulting conditional probability tables are shown in figure \ref{fig:asia_CPTs}. These CPTs fully define the joint distribution of the Bayesian network with 20 parameters. 

%The process of inference in Bayesian networks comes down to calculating the posterior probabilities of the form $P(X|\mathcal{E}=e)$ from the joint probability, which is defined as the product of all conditional probability distributions. 
We use the technique of variable elimination to perform exact inference on our Bayesian networks. There are some cases where this method fails because the query contains some evidence combination it has never seen before. %For example, when the ``asia'' node is always observed as ``no'' in the training set (which is probable for small sizes, since its probability to be ``yes'' is only 0.01), the conditional probability table does not provide any estimate for $P(asia = yes)$, and so it cannot readily answer queries of the form $P(X \mid asia = yes, \mathcal{E} = e)$. 
When coming across such a case during test time, we simply throw out the evidence for this particular query and take $P(X)$ as an estimate. We believe this makes for a fairer comparison than simply ignoring those queries, since our NN is in fact able to provide an estimate for all queries, even if it has never seen a particular evidence set before. 

\subsection{Neural network training details} \label{app:NN_training_details}

Our neural network architecture and its training procedure are implemented in \textit{Pytorch} \citep{pytorch}.

As shown in Fig.~\ref{fig:NN_training}, our neural network is made up of 3 layers. The network receives an input vector of dimension $k$. In our ``Asia'' example, $k$ is equal to 14 (all 7 variables have 2 classes). First, the values of the target variables (as selected by the evidence mask) are substituted by their respective value from $v_0$. The full input is then transformed to dimension $h$, using an input-to-hidden linear layer with tanh activation. Then, a hidden-to-hidden linear layer of size $h$, again with tanh activation is applied. %  layer applies another transformation, but retains dimension $h$ (also using the Tanh activation function). 
Finally, the hidden-to-output layer transforms the activations back to dimension $k$. This layer applies $N$ softmax functions to transform the activations belonging to each variable separately into normalised probability values, as depicted in Fig.~\ref{fig:NN_training}.

For initialisation of the input vector on the positions of the target variables, we use the vector $v_0$ of size $k$, obtained by applying the hidden-to-output layer to a trainable vector of size $h$, followed by the per-variable softmax normalisation. When no evidence is present, $v_0$ serves as the full input vector, and leads to the model predicting the empirical mean probability for all variables.
%a trainable parameter vector of dimension $h$. This embedding is first transformed into a vector of dimension $k$ using the hidden-to-output layer. This layer returns a probability distribution per variable, due to the application of the softmax. These initialisation values are then substituted in the input vector on the corresponding positions of the target variables. 
% We can interpret these values as the network's prior beliefs on the distribution of each variable when no evidence is provided. 

We chose a hidden size $h$ of 50, since we noticed this allowed enough flexibility while still constraining the parameter space sufficiently to avoid overfitting. With these dimensions, our network has 4014 trainable parameters in total. We use the Adam optimiser \citep{adam_optimizer} with a learning rate of 0.001 and otherwise default parameters in Pytorch. The training samples are fed to the neural model in batches of size 16, and the model is trained for 500 epochs. %, without early stopping. 
In the NN+REG model, we additionally use a regularisation batch size of 16. %, along with each equally sized batch for , so every training sample sees one randomly sampled independence relation.
%We use a batched approach to calculate the regularisation loss $\mathcal{L}^{\text{REG}}$. This is mostly for efficiency purposes, since looping over 191 independence relations after every batch of 16 training samples would result in very slow training. 
%We used a regularisation batch size of 16, the same as the training batch size. 
%For every training sample that contributes to the loss, we also test one independence relation and let that contribute to the regularisation loss. 

The hyperparameter $\alpha$ controls the trade-off between training and regularisation loss. The MSE naturally has a smaller order of magnitude than the cross-entropy loss, therefore rather large choices for $\alpha$ (10, 100, 1000) work best. We settled for $\alpha$ equal to 10 since this seemed to lead to the best performance of the NN+REG method, though we did not observe a big difference between any choice of $\alpha$ within the range of 1 to 1000.

\end{document}